\documentclass[10pt,twocolumn,letterpaper]{article}

\usepackage[pagenumbers]{cvpr} 

\definecolor{cvprblue}{rgb}{0.21,0.49,0.74}
\usepackage[pagebackref,breaklinks,colorlinks,allcolors=cvprblue]{hyperref}
\usepackage{multirow}
\usepackage{xcolor}

\newif\ifcomments
\commentstrue  

\def\paperID{3565} 
\def\confName{CVPR}
\def\confYear{2025}
\usepackage{cuted}
\usepackage{tabularx}

\usepackage{overpic}

\title{OphCLIP: Hierarchical Retrieval-Augmented Learning for Ophthalmic Surgical Video-Language Pretraining}

\author{Ming Hu$^{1,2\ast}$, Kun Yuan$^{1,3,4\ast}$, Yaling Shen$^{2\ast}$, Feilong Tang$^{1,2}$, Xiaohao Xu$^5$, Lin Zhou$^1$, \\Wei Li$^{1,7}$, Ying Chen$^{1,8}$, Zhongxing Xu$^2$, Zelin Peng$^7$, Siyuan Yan$^2$, Vinkle Srivastav$^3$, Diping Song$^1$, \\Tianbin Li$^1$, Danli Shi$^6$, Jin Ye$^{1,2}$, Nicolas Padoy$^3$, Nassir Navab$^4$, Junjun He$^{1\dagger}$, Zongyuan Ge$^{2\dagger}$ \\
$^1$Shanghai AI Laboratory, 
$^2$Monash University,
$^3$University of Strasbourg \\
$^4$Technische Universit\"at M\"unchen,
$^5$University of Michigan, Ann Arbor \\
$^6$The Hong Kong Polytechnic University,
$^7$Shanghai Jiao Tong University,
$^8$Xiamen University
}

\begin{document}
\maketitle

\let\thefootnote\relax\footnotetext{\noindent$^\ast$Equal contribution $^\dagger$Corresponding author}
\let\thefootnote\relax\footnotetext{$^\bigstar$Project Page: \href{https://github.com/minghu0830/OphCLIP}{\textit{https://github.com/minghu0830/OphCLIP}}}



\begin{abstract}



Surgical practice involves complex visual interpretation, procedural skills, and advanced medical knowledge, making surgical vision-language pretraining (VLP) particularly challenging due to this complexity and the limited availability of annotated data. To address the gap, we propose OphCLIP, a hierarchical retrieval-augmented vision-language pretraining framework specifically designed for ophthalmic surgical workflow understanding. OphCLIP leverages the OphVL dataset we constructed, a large-scale and comprehensive collection of over 375K hierarchically structured video-text pairs with tens of thousands of different combinations of attributes (surgeries, phases/operations/actions, instruments, medications, as well as more advanced aspects like the causes of eye diseases, surgical objectives, and postoperative recovery recommendations, etc). These hierarchical video-text correspondences enable OphCLIP to learn both fine-grained and long-term visual representations by aligning short video clips with detailed narrative descriptions and full videos with structured titles, capturing intricate surgical details and high-level procedural insights, respectively. Our OphCLIP also designs a retrieval-augmented pretraining framework to leverage the underexplored large-scale silent surgical procedure videos, automatically retrieving semantically relevant content to enhance the representation learning of narrative videos. Evaluation across 11 datasets for phase recognition and multi-instrument identification shows OphCLIP's robust generalization and superior performance. 
\end{abstract}
\vspace{-0.5cm}    
\section{Introduction}
\label{sec:intro}

\begin{figure}[t!]
\centering
\begin{subfigure}[t]{\linewidth}
    \centering
    \includegraphics[width=0.85\linewidth]{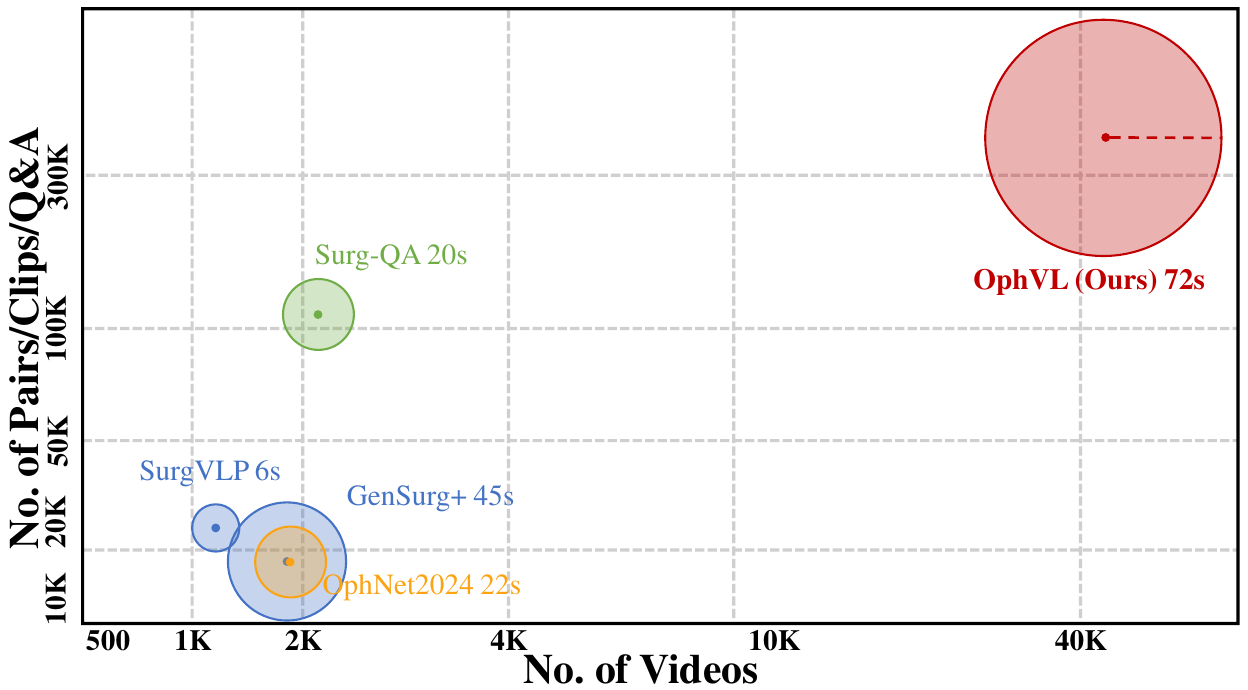}
    \label{data1}
\end{subfigure}

\begin{subfigure}[t]{\linewidth}
    \centering
    \includegraphics[width=0.85\linewidth]{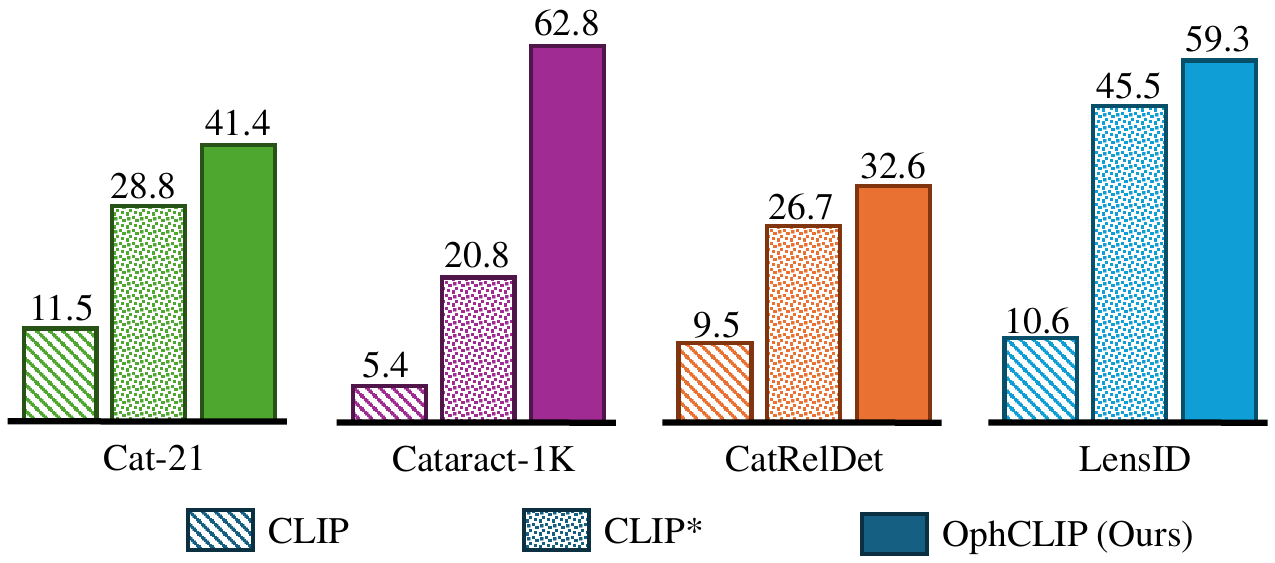} 
    \label{data2}
\end{subfigure}
\caption{\textbf{Dataset comparison and results comparison.} \textbf{TOP:} comparison of our OphVL with existing fully-supervised learning (FSL), VLP, and Q\&A datasets. \textbf{Bottom:} accuracy comparison of CLIP, CLIP* (CLIP fine-tuned on OphVL dataset), and OphCLIP (ours) on phase recognition datasets.}
\vspace{-0.5cm}
\label{fig:fig1}
\end{figure}

The advancement of surgical AI offers opportunities to enhance surgeons' capabilities and revolutionize surgical workflows~\cite{pakkasjarvi2023artificial, varghese2024artificial, schmidgall2024gp}. A key objective is to develop a generalist multi-modal system that can comprehend diverse surgical scenes and communicate with medical professionals using natural language~\cite{schmidgall2024general}, enhancing preoperative planning, intraoperative assistance, and postoperative management. While surgical data science has made progress in uni-modal tasks like surgical phase recognition~\cite{liu2023skit, SchoeffmannHKPM18, wagner2023comparative, hu2024ophnet}, instrument segmentation~\cite{zhou2023text, shvets2018automatic, ceron2022real, laina2017concurrent, pakhomov2019deep, pakhomov2020towards}, and lesion detection~\cite{silva2014toward, bernal2015wm, Smedsrud2021, jha2021comprehensive, mohapatra2022upolyseg}, the exploration of multi-modal representation learning remains limited. 

In general computer vision, models like CLIP~\cite{radford2021learning} have shown success in understanding visual concepts through natural language supervision~\cite{li2023unmasked, luddecke2022image}, enabling a shift from task-specific to generalist models~\cite{ni2022expanding, zou2023generalized, zou2024segment} with less downstream task-specific fine-tuning~\cite{zhou2018end, wang2018reconstruction}. However, surgical multi-modal representation learning poses unique challenges, including specialized medical terminology, and limited data availability, making large-scale datasets difficult to achieve. These challenges, along with the need for expert annotations and the complexity of surgical video data, which often span several hours and involve domain-specific activities within a confined field of view, highlight the gap between general computer vision datasets~\cite{miech2019howto100m, radford2021learning, bain2021frozen} and surgical datasets~\cite{hu2024ophnet, yuan2023learning, honarmand2024vidlpro}, as shown in Fig.~\ref{fig:fig1}, demanding an innovative solution for surgical multi-modal representation learning.

In this work, we develop OphVL, the first large-scale VLP dataset for ophthalmic surgery understanding, featuring 375K carefully processed video-text pairs from 7.5K hours of narrative open-source surgical videos. These videos cover a diverse set of attributes, including surgeries, phases, instruments, medications, and advanced aspects like eye disease causes, surgical objectives, and postoperative recovery recommendations. Unlike previous datasets with noisy and weakly aligned video-text pairs~\cite{miech19howto100m}, OphVL applies extensive data processing and large language models (LLMs) to align the video-text pairs and enrich the textual information. Specifically, we define a set of essential concepts (surgery type, operation type, instrument, medication, anatomy, and surgical objectives) and use LLMs to refine narrative texts to focus on these concepts, enriching the textual information at observational and reasoning levels. We then segment video clips and pair them with narrations, while linking full videos to title summaries to create hierarchical video-text pairs at both clip and video levels. Most previous datasets focus only on narrative videos~\cite{miech19howto100m,bain2021frozen} and overlook the potential of large-scale silent surgical videos. Therefore, our OphVL also includes 30K silent videos as an additional knowledge base for the following pretraining. 


We introduce OphCLIP, a hierarchical retrieval-augmented VLP framework for ophthalmic video-language pretraining, designed to leverage hierarchical video-text pairs from the OphVL dataset. This approach is inspired by how surgeons often break down the understanding of long surgical videos into shorter activity segments, using this granular knowledge to inform their grasp of silent procedural videos. Surgeons often begin by studying narrated surgical videos to learn specific techniques, building foundational knowledge through the sequence of narrations and the title summaries. When transitioning to silent videos with similar content, they leverage these learned representations to effectively transfer the knowledge. 

Similarly, OphCLIP learns short- and long-term visual representations by pairing short video clips with detailed narrative texts for fine-grained features and longer video segments with high-level title summaries for broader context. This approach captures procedural flow and high-level decision-making by integrating both fine- and coarse-grained information. We also introduce retrieval-based augmentation during the pretraining, by constructing a dynamic knowledge base from large-scale silent surgical videos, storing visual and textual embeddings. By retrieving semantically similar silent videos, we add them as auxiliary supervisory signals, facilitating knowledge transfer across narrative and silent procedure videos. OphCLIP integrates these signals into its pretraining framework, learning robust, transferable representations for diverse ophthalmic procedures and achieving state-of-the-art zero-shot performance, as shown in Fig.~\ref{fig:fig1}. Our contributions are as follows: 

\begin{itemize}[leftmargin=*] \item[$\bullet$] \textbf{OphVL Dataset:} We constructed OphVL, the first large-scale and most comprehensive VLP dataset for ophthalmic surgical understanding to date, comprising 375K clip-text pairs with a total duration of 7.5K hours, which is 15$\times$ larger than the current largest surgical VLP dataset. OphVL includes tens of thousands of different combinations of attributes, such as surgeries, phases/actions, instruments, medications, as well as more advanced aspects like the causes of eye diseases, surgical objectives, and postoperative recovery recommendations.


\item[$\bullet$] \textbf{OphCLIP:} We present OphCLIP, a hierarchical ophthalmic surgical VLP framework that aligns short video clips with narrative texts and full videos with high-level summaries, enhancing both fine-grained and long-term visual representation learning. By incorporating silent videos using retrieval-based supervision, OphCLIP enriches its video understanding and learns robust representations for diverse surgical procedures.

\item[$\bullet$] \textbf{Comprehensive Zero-shot Evaluation:} We conduct extensive evaluations and ablation studies of OphCLIP across 11 datasets (sub-datasets) in two tasks: phase and multi-instrument recognition, demonstrating strong generalizationability across tasks of varying granularities. \end{itemize}

\section{Related Work}
\label{sec:related_work}

\begin{table*}[t!]
    \large
    \centering
\resizebox{.98\textwidth}{!}{
    \begin{tabular}{c|cccccccc}
    \toprule
    \textbf{Datasets}  & \textbf{Modality} & \textbf{Source}  & \textbf{Pairs Generation} & \textbf{Text } & \textbf{Videos} & \textbf{Images} & \textbf{Pairs/QA/Clips}  &  \textbf{Avg. Clip Length}   \\
    \midrule
    HowTo100M~\cite{miech19howto100m} & Natural & Self-built & Automatic & Caption & 1.2M &  - & 136M & 4s \\
    ANet-Captions~\cite{krishna2017dense} & Natural & Public & Manual & Caption & 20K & - & 100K & - \\
    Endo-FM~\cite{wang2023foundation}  & Endoscope & Public\&Private & - & - & - & 500M & 33K & 5s \\ 
    SurgVLP~\cite{yuan2023learning}  & Endoscope & Self-built & Automatic & Caption & 1,326 & - & 25K & 6s \\
    GenSurg+~\cite{honarmand2024vidlpro}  & Endoscope & Public & Automatic & Caption & 1.8K & - & 17K & 45s \\
    Surg-QA~\cite{li2024llava} & Endoscope & Public\&Self-built & Automatic & QA & 2,151 & - & 102K  & 20s \\
    GP-VLS~\cite{schmidgall2024gp} & Endoscope & Public & Automatic & QA & - & - & 120K & - \\
    OphNet2024~\cite{hu2024ophnet} & Ophthalmic Scope & Self-built & Manual & FSL & 1,969 & - & 17,508/14,674 & 18s/22s \\
    \midrule
    OphVL (Ours) & Ophthalmic Scope & Self-built & Semi-automatic & Caption & 44,290 & 960M & 375K & 72s \\
    \bottomrule
    \end{tabular} 
    }
\caption{\textbf{Comparison of OphVL with existing FSL, self-supervised pre-training, VLP, and Q\&A datasets in natural and surgical modalities.} OphVL encompasses larger-scale video-text pairs, which is 15$\times$ larger than the current largest surgical VLP dataset. For OphNet2024, ``*/*'' denotes operation/phase level. ``Public'' denotes that the data comes from open-source datasets, while ``Self-built'' denotes that the dataset has been newly collected and organized.}
\vspace{-0.5cm}
\label{tab:comp_general}
\end{table*}

\noindent \textbf{Surgical Vision-Language Pretraining.}
Recent research on deep learning for surgical applications has primarily focused on uni-modal tasks. However, it has largely overlooked advancements in next-generation vision-language models (VLMs)~\cite{seenivasan2023surgicalgpt, wang2024surgical, yuan2024procedure} and GPT frameworks~\cite{wang2023clinicalgpt, zhang2023huatuogpt, nori2023capabilities, qin2022medical}, which enables broader computer vision applications. Many studies have demonstrated the efficacy of utilizing natural language supervision from aligned text to acquire robust visual representations~\cite{bain2021frozen, yuan2021multimodal}. These methods typically leverage contrastive learning~\cite{oord2018representation} to associate videos (or images) with their corresponding narrations (or captions). However, they face sample efficiency challenges with surgical VLP datasets due to noisy transcriptions, limited variability in phase-level descriptions, and strong temporal dependencies in surgical procedures. Recent work has improved data efficiency and zero-shot performance in CLIP-like models through techniques such as text augmentation via EDA~\cite{li2021supervision}, masked token modeling~\cite{sun2019videobert}, captioning loss~\cite{yu2022contrastive}, and knowledge-based, hierarchical-aware augmentations~\cite{yuan2024hecvl, zhang2018cross}. However, the limited scale of surgical video-language pretraining datasets and unique challenges of surgical videos, such as extended durations, narrow fields of view, and procedural variability, continue to restrict the development and evaluation of surgical VLMs.

\noindent \textbf{Video Data for Self-Supervision.}
Recent studies increasingly leverage video data for enhancing self-supervised learning in VLMs~\cite{schiappa2023self, kumar2023large, kwon2024self}. Large datasets from millions of publicly available YouTube videos support training across diverse scenarios, and localized narratives provide dense, frame-level annotations that benefit both single-image and temporal tasks. In medical applications, similar efforts have been directed toward constructing large-scale multimodal datasets, utilizing sources like hospital-based radiological reports~\cite{johnson2019mimic, chen2024chexagent} and publicly accessible platforms such as YouTube and Twitter~\cite{huang2023visual, ikezogwo2024quilt, yuan2023learning, ming2023nurvid, Chen_2023_CVPR} for vision-language pretraining. Although video data introduces noise from varying quality and unfiltered content, advancements in automatic speech recognition (ASR) mitigate this issue by enabling large-scale extraction of cleaner text data directly from audio tracks, improving dataset relevance and model reliability~\cite{ikezogwo2024quilt}. Large language models (LLMs) further aid annotation by offering context-aware insights, reducing manual labeling needs, and enhancing coherence. 

\noindent \textbf{Retrieval-augmented Models.} In the NLP field, retrieval-augmented language models leverage external knowledge to boost performance across tasks~\cite{borgeaud2022improving, ram2023context}. Similarly, recent advancements in the vision-language domain retrieve semantically related samples to enhance tasks, such as image recognition~\cite{long2022retrieval, iscen2023retrieval}, captioning~\cite{li2024evcap, ramos2023smallcap, sarto2022retrieval, xu2024retrieval}, and knowledge-based visual question answering~\cite{schwenk2022okvqa}. While vision-language pretraining~\cite{hu2023reveal} has adopted retrieval-augmentation, with the CLIP model for cross-modal retrieval, it mainly targets images and is suboptimal for surgical video data, particularly in linking narrative and silent procedural videos. In contrast, our OphCLIP builds a dynamically updated knowledge base with video titles as queries for efficient cross-video retrieval and pretraining.
\section{Data Engine }
\label{sec:data_engine}

\begin{figure*}[t!]
\centering
\includegraphics[width=0.95\linewidth]{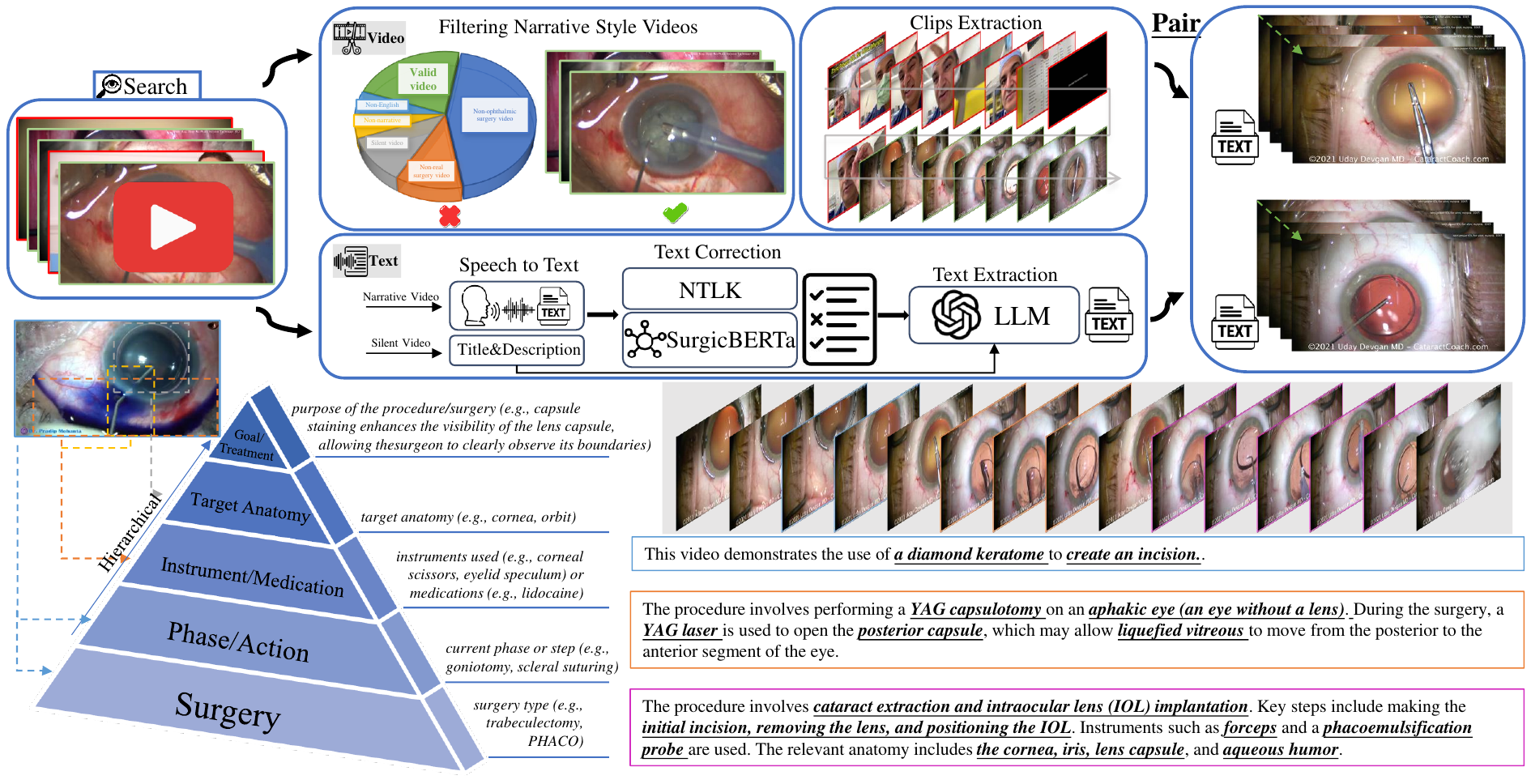}
\caption{\textbf{Overview of OphVL construction pipeline.} The curation pipeline starts with collecting real-world ophthalmic surgery videos and channels using over 3K expert-identified keywords. Next, we filter videos based on their “narrative style” to ensure rich explanatory content. For text extraction, we use ASR models to transcribe audio, followed by denoising and quality control using NTLK and SurgicBERTa to refine and correct medical terminology. Post-processing is done using LLMs to extract structured surgical descriptions.}
\vspace{-0.5cm}
\label{data}
\end{figure*}

We develop OphVL, a dataset of large-scale ophthalmic surgery video-text pairs for VLP. OphVL curation pipeline comprises the following steps: (1) collecting real-world ophthalmic surgery channel and video data, (2) filtering ophthalmic surgery videos based on a specific ``narrative style'', (3) extracting and denoising video segments and narration texts from videos using various models, tools, and algorithms, (4) rewriting narration texts via LLMs to focus on essential surgery-specific concepts, and finally, (5) extracting frames from video clips to construct video-text pairs.


\subsection{Collecting Clip and Text Pairs from YouTube}
\label{collection}
\noindent \textbf{Collecting Representative Channels and Videos.}
In collaboration with three practicing ophthalmologists, we compile a comprehensive list of over 3K terms relevant to ophthalmic surgery, derived from extensive literature review. These terms encompass but are not limited to, surgical names, procedural steps, instrument usage, medications, and postoperative complications. Using these keywords, we manually search YouTube to identify ophthalmic surgery channels. Based on our experience, channel-based searches yield a more concentrated and higher-quality collection of ophthalmic surgical videos than individual keyword searches. Ultimately, we identify the YouTube channel IDs for 410 ophthalmic surgery videos and download them in bulk. During the download process, we prioritize the highest-resolution versions, filtering out videos shorter than 30 seconds or with resolution below 224p, resulting in a collection of approximately 100K videos.

\noindent \textbf{Filtering for Narrative-Style Surgical Videos.}
In this step, we assess each video from each channel to determine (1) whether it depicts real ophthalmic surgeries or contains usable surgery segments, and (2) whether it qualifies as a narrated video with rich explanatory voiceover. For (1), we identify relevant videos by extracting keyframes, which are automatically generated using PySceneDetect to mark the start or end of scenes with significant visual changes. We then train and apply a ResNet50 image classifier to determine if the keyframes are microscope images of ophthalmic surgeries. Videos with over 80\% of keyframes classified as ophthalmic microscope images are labeled as valid videos. For (2), we use 
it to detect the proportion of human voice in the video, setting a threshold of 0.2. Videos below this threshold are flagged as silent or lacking sufficient explanatory narration. For these videos, we collect their titles and clip metadata to construct a knowledge base, which we use to enhance representation learning.

\subsection{Text Extraction using ASR and Text Denoising}
To tackle the challenges of ASR with medical terminology in YouTube captions, we employed the large-scale open-source Whisper Large-V3 model~\cite{radford2023robust} for speech-to-text conversion by directly transcribing entire speech segments. We then developed a transcription denoising and quality control pipeline consisting of: (i) applying the Rake keyword extraction algorithm to identify key phrases (up to four words) and refining them by removing stopwords [52]; (ii) using SurgicBERTa~\cite{SurgicBERTa2023}, a language model pre-trained on surgical texts, to validate and correct each refined entry for alignment with known surgical terminology and context; (iii) conditioning a large language model with example prompts to correct spelling errors within sentence context; and (iv) prompting the language model to provide structured summaries of the captions, focusing on key components such as surgery type, phase/action, instrument, medication, anatomical target, and procedure goal.

\subsection{Aligning Clip/Image and Text Pair}
Due to frequent sentence segmentation discontinuity in Whisper transcriptions—where coherent sentences are often split across multiple timestamps, we develop a heuristic algorithm that merges timestamps based on punctuation and linking words, ensuring semantic continuity and improving GPT-4o summaries. For clip extraction, we align segments with subtitle timestamps. For silent videos, the classifier (Sec.~\ref{collection}) samples frames at 1 FPS to extract surgical clips. Titles and metadata are collected to build a knowledge base, enhancing representation learning. Finally, for all clip-text pairs, frames are extracted at 0.5 FPS for pre-training.

\subsection{OphVL Statistics}
As shown in Tab.~\ref{tab:comp_general}, the final OphVL dataset comprises 375,198 clip-text pairs extracted from 13,654 narrated videos and 30,636 silent videos (totaling 9363 hours). On average, the clips have a duration of 72 seconds and a resolution of 1500$\times$912, with over 65\% of the videos having a resolution equal to or greater than 1280$\times$720. According to our rough estimation, our textual concepts include tens of thousands of different combinations of attributes, such as surgeries, surgical phases/operations/actions, surgical instruments, medications, as well as more advanced aspects like the causes of eye diseases, surgical objectives, and postoperative recovery recommendations. Please refer to the supplementary for more dataset statistical details.





\section{OphCLIP}
\label{sec:ophclip}

We introduce OphCLIP, the hierarchical retrieval-augmented video-language pretraining framework, including hierarchical video-text correspondences (Sec.~\ref{hier_correspondence}), our contrastive learning approach for fine- and coarse-grained representations (Sec.~\ref{hier_pretraining}), and our strategy for leveraging silent videos as a knowledge base to enhance representation learning (Sec.~\ref{KB}).

\begin{figure*}[t!]
\centering
\begin{overpic}[width=\textwidth]{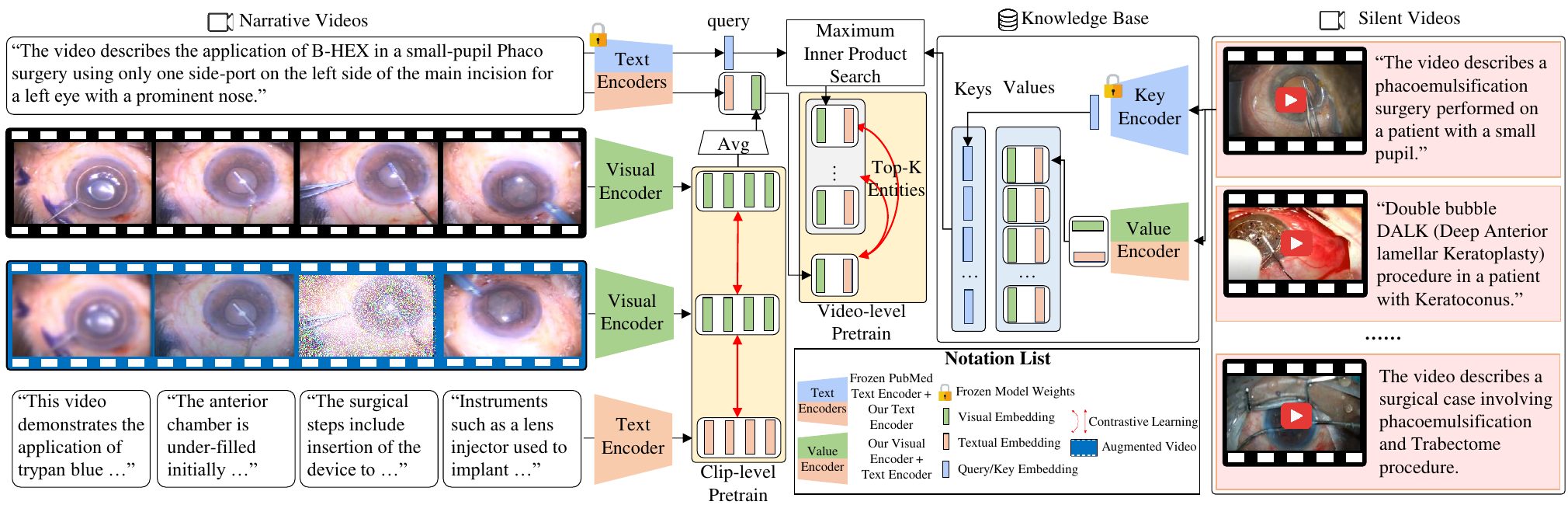}
\end{overpic}
\caption{\textbf{OphCLIP's framework for video-language pretraining.} OphCLIP performs vision-language pretraining at both clip and video levels, learning short-term visual representations from narrations and long-term representations from titles, enhanced by a knowledge base. OphCLIP has several components: Narrative videos with associated narrative texts are processed through visual and text encoders, creating clip-level multi-modal embeddings; Silent videos' multi-modal embeddings are stored in the dynamically updated memory bank, constructing the knowledge base; Video-level pretraining uses maximum inner product search to retrieve relevant top-K silent videos' embeddings based on queries to enhance the video-level pretraining.}
\vspace{-0.5cm}
\label{arch}
\end{figure*}

\subsection{Hierarchically Video-text Correspondences}
\label{hier_correspondence}

We leverage our curated OphVL dataset to train OphCLIP. The OphVL dataset, illustrated in Fig.~\ref{arch}, is a partially hierarchically annotated video collection, denoted as $\mathcal{D} = \{(V^n_i, N^n_i, T^n_i)\}_{i=1}^{|\mathcal{D}_N|} \cup \{(V^s_i, T^s_i)\}_{i=1}^{|\mathcal{D}_S|}$. Here, $\mathcal{D}_N$ contains narrative videos ($V^n_i$) paired with clip-level narrations ($N^n_i$) that describe surgical observations and reasoning, as well as high-level video summaries ($T^n_i$). In contrast, $\mathcal{D}_S$ comprises silent videos ($V^s_i$) paired only with video-level summaries ($T^s_i$). Each video, whether narrative or silent, is segmented into clips, represented as $V^{n/s}_i = \{v^{n/s}_{ij}\}_{j=1}^{|V^{n/s}_i|}$, with each clip $v_{ij}$ providing a visual counterpart to the narration $n_{ij}$ in narrative videos. For instance, a clip’s narration describes ``The anterior chamber is under-filled...'' while the high-level title summarizes the procedure as ``Application of B-HEX in a small-pupil Phaco surgery...''. This hierarchical text structure allows the model to capture both fine-grained surgical details at the clip level and overall procedural goals at the video level, enabling robust vision-language representations across diverse granularity levels.

\subsection{Hierarchical Vision-language Pretraining}
\label{hier_pretraining}


OphCLIP has two pretraining stages: clip- and video-level pretraining to learn fine, short-term, and coarse, long-term representations, respectively. OphCLIP adopts CLIP-like architecture~\citep{radford2021learning}, using visual and textual encoders, $f_v$ and $f_t$, to generate embeddings for frames and texts. A single set of visual and textual encoders is used across both pretraining stages. For clip-level pretraining, each video clip $v^n_{ij}$ is paired with its narration $n_{ij}$ to learn short-term representations, denoted as $f_v(v^n_{ij})$ and $f_t(n_{ij})$. At video level, the entire video $V_i$ (including both narrative and silent types) is paired with a summary text $T_i$ to form long-term features, represented by $f_v(V_i)$ and $f_t(T_i)$. This high-level summary $f_t(T_i)$ captures the overall semantic context, supporting deeper long-term reasoning within visual representations.

\noindent \textbf{Clip-level Pretraining.} For OphCLIP’s clip-level pretraining, we use an InfoNCE loss~\citep{oord2018representation} to align short-term video clips $v_{ij}$ with their corresponding narration texts $n_{ij}$. This objective maximizes similarity between visual features $f_v(v_{ij})$ and textual features $f_t(n_{ij})$, as shown below in Eq.~\ref{eq:loss_clip_vl}:
\begin{equation}
        L^{\text{vl}}_{\text{clip}} = \frac{1}{B} \sum_{i=1}^B \log \frac{\exp(f_v(v_{ij})^\top f_t(n_{ij}))}{\sum_{k=1}^B \exp(f_v(v_{ij})^\top f_t(n_{kj}))},
    \label{eq:loss_clip_vl}
\end{equation}
where $B$ is the batch size. Positive pairs consist of temporally aligned video-text pairs, while other pairs in the batch serve as negatives, enabling OphCLIP to learn from short-term video-text correspondences.

To further refine visual features, we incorporate SimSiam self-supervision~\citep{chen2021exploring}. By applying random augmentations to create two views of each video clip, we maximize similarity within positive pairs, formalized as $L^{\text{vv}}_{\text{clip}}$:
\begin{equation}
\hspace{-2mm}
    \scalebox{0.99}{
    $L^{\text{vv}}_{\text{clip}} = \frac{1}{B} \sum_{i=1}^B \log \frac{\exp(f_v(v_{ij})^\top f_t(n_{ij}))}{\sum_{k=1}^B \exp(f_v(v_{ij})^\top f_v(Aug(v_{kj})))}.$
    }
    \label{eq:loss_clip_vv}
\end{equation}

The combined clip-level objective, $L_{\text{clip}} = L^{\text{vl}}_{\text{clip}} + L^{\text{vv}}_{\text{clip}}$, strengthens fine-grained visual-textual alignment.

\noindent \textbf{Video-level Pretraining.} At the video level, we aim to capture long-term procedural context by aligning each narrative video’s high-level title summary $T$ with a sequence of video segments $V_i = \{v_{i1},...v_{i|V_i|}\}$. This process uses the following loss:
\begin{equation}
        L^{\text{narrative}}_{\text{video}} = \frac{1}{B} \sum_{i=1}^B \log \frac{\exp(f_v(V_i)^\top f_v(T_i))}{\sum_{k=1}^B \exp(f_v(V_i)^\top f_t(T_k))}.
\end{equation}

This objective aligns the entire video representation $f_v(V_i)$ with its summary $f_t(T_i)$ while treating summaries from other videos as negatives. To efficiently obtain long-term visual representations for whole procedure videos, we aggregate short-term video clip features using average pooling. This approach ensures computational efficiency and preserves frame-wise visual features, represented as $f_v(V_i) = Agg\left({f_v(v_{ij})}_{j=1}^{|V_i|}\right)$.




\subsection{Silent Videos as Knowledge Pool} 
\label{KB}
In addition to learning representations from the narrative videos, our OphCLIP explores silent surgical videos to form a contextual knowledge base that facilitates knowledge transfer and enriches multi-modal representations, as shown in Fig.~\ref{arch}. Using the title of a narrative video as a query, the retriever matches it to relevant silent videos stored in the memory bank, enhancing OphCLIP’s long-term visual representations by integrating additional procedural context from these silent videos.

\noindent \textbf{Query Encoding.} To retrieve relevant information from silent videos, we first encode the titles of narrative videos as query embeddings. Specifically, we use the PubMedBERT~\citep{pubmedbert} as the query encoder $f_{\text{query}}$ to transform the title text into a high-dimensional embedding:
\begin{equation}
    q^n = f_{\text{query}}(T^n).
    \label{eq:query_encoding}
\end{equation}

This query embedding $q$, based on the video’s title, captures its semantic essence and is used to find related content in a silent video pool. Encoding the title text into embeddings allows efficient retrieval within the text modality, avoiding inaccuracies from CLIP’s limited visual understanding of ophthalmic videos.

\noindent \textbf{Memory Bank.} The memory bank module stores multi-modal representations of silent videos. Each silent video $\hat{V^s_k}$ is encoded by the visual encoder $f_v$, and its corresponding title text $\hat{T^s_k}$ is encoded by the text encoder $f_t$ and above-mentioned query encoder $f_{query}$. We use title embeddings as keys in the memory bank, with values comprising visual and textual embeddings of ophthalmic videos and corresponding texts:
\begin{equation} 
\scalebox{0.82}{
    $\text{Memory} = { (\underbrace{f_{\text{query}}(\hat{T^s_k})}_{\text{Key}}, \underbrace{(f_t(\hat{T^s_k}), f_v(\hat{V^s_k}))}_{\text{Values}}) \mid k = 1, \dots, |\mathcal{D}_S| ,} $
}
\label{eq:memory_bank} 
\end{equation}
where $|\mathcal{D}_S|$ represents the number of silent videos in the OphVLP dataset. In our memory bank, both visual and textual representations are stored and updated dynamically to improve hierarchical vision-language pretraining.

\begin{table*}[t]
    \centering
\scalebox{0.95}{
        \begin{tabular}
        {@{\hspace{2pt}}l@{\hspace{5pt}}l@{\hspace{10pt}}c@{\hspace{10pt}}c@{\hspace{5pt}}c@{\hspace{5pt}}c@{\hspace{5pt}}c@{\hspace{5pt}}c}
        
        \toprule[1.1pt]

        \multirow{3}{*}{Task} & \multirow{3}{*}{Dataset} & CLIP~\citep{radford2021learning} & SLIP~\citep{mu2022slip} & LaCLIP~\citep{fan2024improving} & CLIP~\citep{radford2021learning} & CLIP~\citep{radford2021learning} & OphCLIP\\
        & & VTIB16 & VITB16 & VITB16 & RN50 & RN50 & RN50 \\
        & & CLIP400M & YFCC100M & LAION-400M & CLIP400M & OphVL & OphVL \\

        \hline
        
        \multirow{6}{*}{\begin{tabular}[l]{@{}l@{}}\textit{Phase} \end{tabular}}

        &Cat-21~\citep{PrimusPTMEBS18} & 11.5 / 2.6 & 7.7 / 2.4 & 9.6 / 3.6 &  13.3 / 2.7 & 28.8 / 17.6 & \textbf{41.4} / \textbf{28.8}  \\ 
        &Cataract-1K~\citep{ghamsarian2024cataract} & 5.4 / 1.6  & 5.7 / 2.8 & 10 / 1.7 & 6.9 / 2.0 & 20.8 / 15.9 & \textbf{62.8 / 48.5} \\ 
        &Cataract-101~\citep{schoeffmann2018cataract} & 9.9 / 4.1 & 7.3 / 2.6 & 9.8 / 2.4  & 10.0 / 3.3 &  36.2 / 25.5  & \textbf{39.3 / 33.7}   \\ 
        &CatRelDet~\citep{hamsarianTPSS20}  & 9.5 / 7.0 & 10.2 / 6.4 & 11.8 / 4.4 & 15.3 / 11.9  &  26.7 / 23.7 & \textbf{32.6 / 34.2} \\ 
        &LensID~\citep{GhamsarianTPSES21} & 10.6 / 6.5 & 10.6 / 6.3 & 25.8 / 17.8 & 22.9 / 16.3 & 45.5 / 32.1 & \textbf{59.3 / 41.0} \\ 
        \hline
        \multirow{2}{*}{\begin{tabular}[l]{@{}l@{}}\textit{Instrument} \end{tabular}} 
        &Cataract-1K~\citep{ghamsarian2024cataract} & 100.0 / 13.9 & 100.0 / 13.9 & 100.0 / 13.9 & 100.0 / 13.9 & 80.8 / 15.3 &  \textbf{45.1 / 21.2} \\ 
        &CatInstSeg~\citep{FoxTS20} & 100.0 / 20.2  & 100.0 / 20.2 & 100.0 / 20.2 & 100.0 / 20.2 & 87.3 / 20.3 & \textbf{51.1 / 28.3} \\

        \bottomrule[1.1pt]
    
        \end{tabular}}
    \captionsetup{width=\textwidth} 
    \caption{\textbf{The comparison to the OpenAI CLIP and CLIP pretrained on our dataset.} We report Accuracy / F1-score for zero-shot surgical phase recognition. We report False Positive Rate (the lower the better) FPR / mAP for zero-shot instrument recognition.
    }
    \vspace{-0.5cm}
    \label{tbl:zero_shot}
    \end{table*}

\begin{table}[t!]
\centering
\scalebox{0.75}{
\begin{tabular}{ccccc}
   \toprule[1.1pt]
    \multirow{2}{*}{Method} & OphNet-O & OphNet-P~\cite{hu2024ophnet} & CaDIS-F & CaDIS-C~\cite{grammatikopoulou2021cadis} \\ 
    \cmidrule(lr){2-5}
    & Acc / F1 & Acc / F1 & FPR / mAP & FPRs / mAP \\ 
    \midrule
    CLIP~\cite{radford2021learning}  & 0.7 / 0.4 & 3.2 / 0.7 & 13.0 / 8.9 & 28.1 / 22.5 \\
    LaCLIP~\cite{fan2024improving} & 1.0 / 0.3 & 3.2 / 0.7 & 13.0 / 8.9 & 30.1 / 22.6 \\
    \midrule
    CLIP*~\cite{radford2021learning}  & 2.5 / 0.8 & 5.0 / 1.6 &  13.1 / 9.3 & 28.5 / 22.6 \\
    OphCLIP~ & \textbf{7.1} / \textbf{2.3} & \textbf{18.2} / \textbf{4.8} & \textbf{14.7} / \textbf{10.6} & \textbf{28.7} / \textbf{23.5} \\
   \bottomrule[1.1pt]
\end{tabular}
}
\caption{\textbf{Fine- vs. Coarse-grained recognition.} The OphNet dataset subsets, OphNet-O and OphNet-P, support operation and phase recognition tasks, respectively. CaDIS-F and CaDIS-C, derived from the CaDIS dataset, offer fine-grained and coarse-grained multi-instrument recognition labels. CLIP* represents the CLIP model pretrained on the OphVL dataset.}
\label{tbl:coarse_fine}
\vspace{-0.25cm}
\end{table}
\noindent \textbf{Retriever.} As shown in Fig.~\ref{arch}, our retriever component leverages the query embedding $q$ to perform a maximum inner product search (MIPS) on the memory keys, identifying the top-$K$ silent videos most relevant to the queried narrative video. Specifically, we compute similarity with $q$ for each key embedding $f_{\text{query}}(\hat{T^s_k})$ in memory, selecting the top-$K$ indices that yield the highest similarity scores:
\begin{equation}
    \text{Retrieved Indices} = \text{argsort}_k \left( q^\top f_{\text{query}}(\hat{T^s_k}) \right)[:K].
    \label{eq:retriever}
\end{equation}

Then, the retrieved video-level multi-modal representations $\{(f_v(\hat{V^s_{k_1}}), f_t(\hat{T^s_{k_1}})), \dots, (f_v(\hat{V^s_{k_K}}, f_t(\hat{T^s_{k_K}}))))\}$ from $K$ most relevant silent videos are considered as positive samples to the queried narrative video. We conduct the contrastive learning as Eq.~\ref{eq:loss_video_kb} shows:
\begin{equation} 
\hspace{-3mm}
    \scalebox{0.70}{
        $L^{\text{silent}}_{\text{video}} = \frac{1}{K} \displaystyle\sum_{j=1}^K \log \frac{\exp(f_v(V_i)^\top f_v(\hat{V_{i_j}))} + \exp(f_v(V_i)^\top f_t(\hat{T_{i_j}}))}{\sum_{m=1}^K \exp(f_v(V_i)^\top f_v(\hat{V_{i_m}))} + \exp(f_v(V_i)^\top f_t(\hat{T_{i_m}}))}.$
    }
\label{eq:loss_video_kb} 
\end{equation}

Thus, the final video-level pretraining loss is:
\begin{equation} 
L_{\text{video}} = L^{\text{narrative}}_\text{{video}} + L^{\text{silent}}_{\text{video}}.
\end{equation}

We leverage the retrieved $K$ similar entities to introduce diverse supervisory signals from additional video-text pairs, facilitating knowledge transfer across narrative and silent procedure videos. Our method captures richer contextual information by pairing each video-level feature of the narrative videos with their title text embedding and the retrieved multi-modal embeddings. This approach not only enhances model robustness but also enables efficient retrieval within million-scale datasets. We employ an alternating training strategy, optimizing $L_{clip}$ for a few batches followed by optimizing $L_{video}$ for a few batches, and repeating this cycle. This strategy optimizes the visual and textual for both short-term and long-term features and also avoids the catastrophic forgetting issue~\cite{ashutosh2023hiervl, yuan2024hecvl}. Please refer to the supplementary for more implementation details.

\label{implement}

\section{Experiments}
\label{sec:experiments}

\begin{table}[t!]
\centering
\scalebox{0.85}{
\begin{tabular}{cccc}
   \toprule[1.1pt]
    \multirow{2}{*}{Method} & \multirow{2}{*}{Data (\%)} & Cat-21~\citep{PrimusPTMEBS18} & Cataract-1K~\citep{ghamsarian2024cataract} \\ 
    \cmidrule(lr){3-4}
    & & Acc / F1 & F1 / mAP \\ 
    \midrule
    CLIP~\cite{radford2021learning}   & 100 & 48.2 / 32.6 & 0.0 / 14.1 \\
    CLIP*~\cite{radford2021learning}  & 100  & 59.4	/ 40.5 & 2.0 / 14.2 \\
    OphCLIP~\cite{mu2022slip}            & 100  & \textbf{72.1} / \textbf{57.9} & \textbf{11.6} / \textbf{18.8} \\
    \midrule
    CLIP~\cite{radford2021learning}   & 10   & 40.1	/ 17.3 & 0.0 / 15.5 \\
    CLIP*~\cite{radford2021learning}  & 10   & 47.6 / 26.6 & 1.0 / 15.8 \\
    OphCLIP~\cite{mu2022slip}            & 10   & \textbf{59.5} / \textbf{41.6} & \textbf{15.6} / \textbf{22.2} \\
   \bottomrule[1.1pt]
\end{tabular}
}
\caption{\textbf{Few/Full-shot linear probing.} Accuracy and F1 scores for Cat-21 and Cataract-1K datasets with different methods using 10\% and 100\% training data.}
\label{tbl:few_full_linear}
\vspace{-0.5cm}
\end{table}

\subsection{Datasets}
To evaluate our approach, we conduct experiments on two downstream tasks, i.e., phase recognition and multi-instrument recognition, using 12 datasets (or sub-datasets).
\textbf{(1) Phase Recognition}: Five datasets are used for this task: \textit{Cat-21}~\cite{PrimusPTMEBS18} (11 classes), \textit{Cataract-1K}~\cite{ghamsarian2024cataract} (12 classes), \textit{Cataract-101}~\cite{schoeffmann2018cataract} (10 classes), \textit{CatRelDet}~\cite{hamsarianTPSS20} (5 classes), \textit{OphNet} (96 phases, 232 operations)~\cite{hu2024ophnet}, and LensID~\cite{GhamsarianTPSES21} (3 classes). The implantation and rest phase recognition task represents a specialized configuration of phase recognition, wherein video frames are labeled with only three-phase categories: pre-implantation, implantation, and post-implantation of the lens. \textit{OphNet} additionally provides both phase and operation labels, offering finer granularity for classification.
\textbf{(2) Multi-instrument Recognition}: We select four datasets for multi-instrument recognition: \textit{Cataract-1K}~\cite{ghamsarian2024cataract} (9 classes), \textit{CatInstSeg}~\cite{FoxTS20} (11 classes), and \textit{CaDIS}~\cite{grammatikopoulou2021cadis} (12 classes, 35 classes). \textit{CaDIS} also provides labels at different levels of granularity.

\subsection{Zero-shot Recognition}

In this section, we demonstrate the zero-shot transfer performance of our pretrained OphCLIP model across various downstream tasks. Following CLIP~\citep{radford2021learning}, we keep the pretrained visual and text encoders fixed and format class labels as sentence prompts for classification.

\begin{figure*}[t!]
\centering
\includegraphics[width=0.93\linewidth]{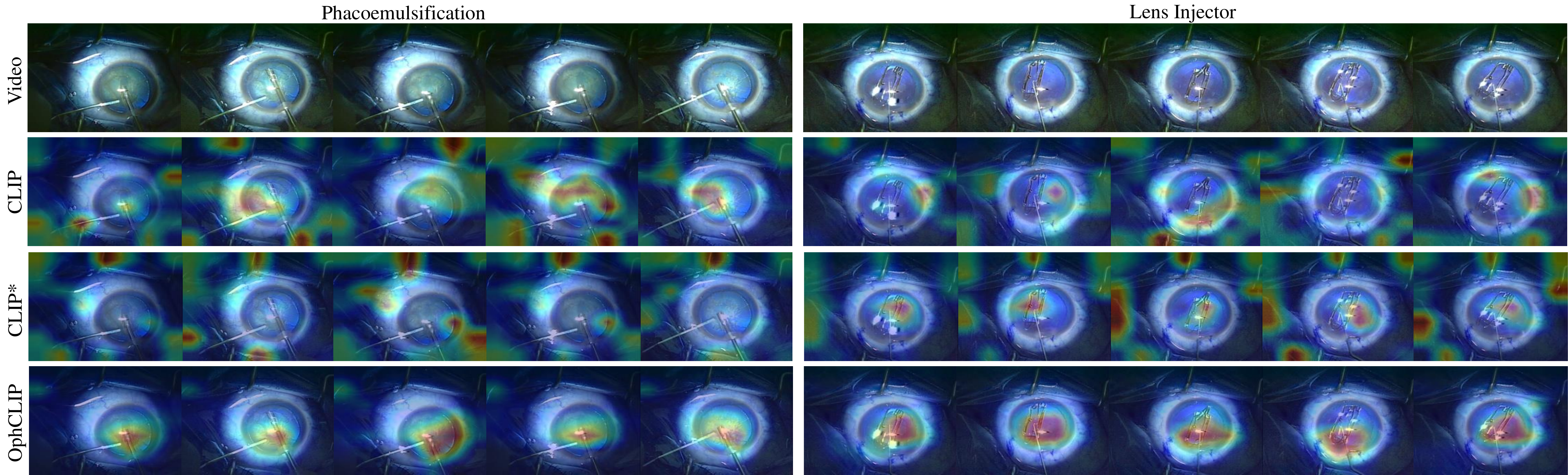}
\caption{\textbf{Attention map visualizations among CLIP, CLIP* (CLIP fine-tuned on OphVL), and OphCLIP (Ours) for phase recognition (left) and instrument recognition (right) examples from unseen Cataract-1K dataset.} \textbf{Left:} For phase recognition (e.g., ``phacoemulsification''), OphCLIP focuses on instruments and anatomy to identify the high-level surgical phase concept. \textbf{Right:} For instrument recognition, pretraining on OphVL enables CLIP* and OphCLIP to attend consistently to domain-specific tools like the lens injector.}
\vspace{-0.5cm}
\label{cam}
\end{figure*}

\noindent \textbf{Phase Recognition.} As shown in Tab.~\ref{tbl:zero_shot}, methods pretrained on OphVL dataset, including CLIP, consistently outperform baselines like vanilla CLIP~\citep{radford2021learning} and SLIP~\citep{mu2022slip} across all surgical phase recognition datasets. This demonstrates the impact of our ophthalmic-specific pretraining dataset. Fig.~\ref{cam} also highlights OphCLIP's capability to understand ophthalmic-specific concepts across both visual and linguistic modalities. The model not only recognizes relevant anatomical and procedural elements in ophthalmic images but also aligns these elements with corresponding medical terminology and context in textual descriptions. This cross-modal understanding enables OphCLIP to focus on regions within the visual data that contribute most significantly to the semantics, demonstrating that the model effectively prioritizes clinically relevant areas.

\begin{table}[t!]
\centering
\scalebox{0.76}{
\begin{tabular}{c|ccc|cc}
\toprule[1.1pt]
{Model}  & {OphVL} & {KB} & {Prompt} & Cat-21~\citep{PrimusPTMEBS18} & Cataract-1K~\cite{ghamsarian2024cataract}  \\
\hline
CLIP~\citep{radford2021learning} & $\times$ & $\times$ & Caption & 13.3 / 2.7 & 6.1 / 2.3 \\
CLIP~\citep{radford2021learning} & $\checkmark$ & $\times$ & Caption & 28.8 / 17.6 & 20.8 / 15.9 \\
OphCLIP & $\checkmark$ & $\times$ & Caption & 34.9 / 22.9 & 60.5 / 44.7 \\
OphCLIP & $\checkmark$ & $\times$ & Mix & 41.3 / 27.7 & 55.9 / 43.0 \\
OphCLIP & $\checkmark$ & $\checkmark$ & Caption & 39.9 / 28.2 & 61.3 / 47.2  \\
OphCLIP & $\checkmark$ & $\checkmark$ & Mix & \textbf{41.4} / \textbf{28.8} & \textbf{62.8} / \textbf{48.5} \\
\bottomrule[1.1pt]
\end{tabular}
}
\caption{\textbf{Ablation study of OphCLIP with various components.} This study evaluates the impact of different OphCLIP components: OphVL (pretraining dataset), KB (knowledge base with silent videos), and Prompt (descriptive phase prompts, comparing basic Caption vs. Mix, which adds keywords to captions). Results are presented as Accuracy / F1-score.}
\label{tab:ablation}
\vspace{-0.5cm}
\end{table}

\noindent \textbf{Multi-instrument Recognition.} 
To perform surgical instrument recognition with pretrained vision-language models, each instrument class is converted into a text prompt. For each input image, we compute cosine similarities between image features and these prompts, generating a similarity score per class. Since instrument recognition is a multi-label task, sigmoid activation is applied to these scores, allowing the model to output independent probabilities for each instrument. As shown in Tab.~\ref{tbl:zero_shot}, baseline models like CLIP and SLIP~\citep{mu2022slip} show high False Positive Rates (FPR) in instrument recognition, with 100\% FPRs on Cataract-1K~\citep{ghamsarian2024cataract} and CatInstSeg~\citep{FoxTS20}, indicating substantial false detections. In contrast, OphCLIP significantly reduces false positives, as shown in Fig.~\ref{cam}. This demonstrates the effectiveness of OphCLIP’s ophthalmic-specific pretraining, enhancing its ability to detect surgical instruments accurately and reduce errors.





\noindent \textbf{Fine-grained vs. Coarse-grained.} 
Tab.~\ref{tbl:coarse_fine} shows that OphCLIP outperforms other models in both phase recognition across different granularities. For OphNet-O (fine-grained) and OphNet-P (coarse-grained), OphCLIP achieves 18.2\% / 4.8\% and 7.1\% / 2.3\% (Acc / F1), significantly surpassing CLIP and LaCLIP~\citep{fan2024improving}. In instrument recognition, OphCLIP achieves an FPR of 14.7\% / mAP of 10.6\% on CaDIS-F and an FPR of 28.7\% / mAP of 23.5\% on CaDIS-C, indicating lower false positives and higher precision. These results confirm OphCLIP’s robust performance and adaptability in tasks of different granularities.


\subsection{Few/Full-shot Linear Probing}

As shown in Tab.~\ref{tbl:few_full_linear}, we evaluate pretrained method's visual encoder on Cat-21 and Cataract-1K using linear probing with both 10\% and 100\% of the data. Our OphCLIP demonstrates substantial performance gains over both CLIP variants, particularly with 100\% of the data. These results show that OphCLIP's visual encoder provides strong, transferable representations for diverse surgical tasks, serving as an effective generalist backbone.

\subsection{Ablation Studies}
We conduct an ablation study on surgical phase downstream datasets to examine the effect of OphCLIP's components (Tab.~\ref{tab:ablation}). Models pretrained with OphVL consistently outperform those without it, showing notable accuracy gains across all tasks. Adding a knowledge base (KB) with more silent videos further boosts performance, evidenced by higher F1 scores on Cataract-1K. Prompt choice also plays a key role in zero-shot phase recognition. The ``Mix'' prompt, which includes keywords like ``instrument,'' ``medication,'' and ``goal'', outperforms the ``Caption'' prompt. This is due to the specialized OphVL corpus focusing on instruments, medications, and procedural goals, enabling better concept capture and improved predictions.




\section{Limitation}
Despite our efforts with OphVL and OphCLIP to enhance perception in ophthalmic surgeries, biases from open-source videos persist, including regional practice variations and inconsistent terminology. Secondly, though OphVL covers a wide range of surgeries, research mainly focuses on cataract procedures due to their prevalence and accessibility, making our validation dataset primarily cataract-based. This limits our ability to validate the model across other surgeries like glaucoma and corneal procedures. While OphCLIP outperforms baseline models in phase recognition and multi-instrument classification, the limited variety restricts evaluation on more complex tasks like anomaly detection and other advanced challenges.

\section{Conclusion}
In this work, we introduce OphCLIP, a specialized vision-language pretraining framework designed for ophthalmic surgery. By constructing the comprehensive OphVL dataset, which includes over 375K clip-text pairs and tens of thousands of ophthalmic surgery-related concepts (surgeries, procedures, instruments, medications, surgical goals, etc.), we enable robust hierarchical learning of both fine-grained and long-term visual representations. Our approach leverages both narrative and silent videos through innovative retrieval-based supervision, resulting in enhanced understanding and generalization across surgical phases and multi-instrument identification tasks. This research sets a new benchmark for ophthalmic surgical workflow understanding and opens avenues for more specialized and context-aware AI applications in ophthalmic surgery.

\clearpage
\setcounter{page}{1}
\maketitlesupplementary

\section{OphVL Dataset}
Fig.~\ref{pair_sample} illustrates the clip-text pair samples we constructed. Through our data processing pipeline, OphVL achieves high-quality modality alignment between ophthalmic surgery videos and descriptive texts.

\section{Experiments}
\subsection{Implementation Details.}

\begin{table}[ht]
\footnotesize
\centering
\begin{tabular}{c|c|c}
\toprule[1.1pt]
\multicolumn{2}{c|}{\textbf{Hyper-parameter}} & \textbf{Value} \\ \hline
\multicolumn{2}{c|}{Epochs} & 60 \\
 \midrule

\multirow{4}{*}{Clip-level Pretraining} & Batch Size & 120 \\
 & Image Size & 224 \\
 & \# of Frames & 8 \\
 & Text Length & 77 \\
 \hline
\multirow{5}{*}{Clip-level Pretraining} & Batch Size & 140 \\
 & Image Size & 224 \\
 & \# of Frames & 8 \\
 & \# of Retrieved Videos & 1 \\
 & Text Length & 77 \\
 \hline
\multirow{4}{*}{Optimization} & Learning Rate & 8e-5 \\
 & Scheduler & Cosine \\
 & Optimizer & Adam \\
 & Momentum & 0.9 \\
 \hline
\multirow{3}{*}{Loss Function} & Temperature & 0.1 \\
& Weight of $L_\text{{clip}}^\text{{vv}}$ & 0.5 \\
& Weight of $L_\text{{clip}}^\text{{vl}}$ & 0.5 \\
\bottomrule
\end{tabular}
\caption{Hyper-parameter details.}
\label{tab:hyperparam}
\end{table}

\noindent \textbf{Architecture.} We use the CLIP-like architecture~\cite{radford2021learning} with two branches, i.e., visual and textual encoders. We use the ResNet-50 as the visual encoder from the ImageNet initialization. We apply BioClinicalBert~\cite{huang2019clinicalbert} as the textual encoder, which is pretrained on the clinical notes. Then we apply the average pooling at the visual features to generate the visual embeddings. We apply a linear projection layer at the end of Bert model's $[CLS]$ token to generate textual embeddings. We use $768$ as the dimensionality of the embedding space. 

\noindent \textbf{Pretraining Setups.}  In total, we use $8$ RTX-$4090$ with $24$ GB and train for $2$ days. We first perform clip-level pretraining for $40$ epochs and then apply the hierarchical pretraining strategy, which alternatively trains with $3$ epochs of clip-level video-text pairs, followed by $2$ epochs of video-level video-text pairs. We use a batch of $120/140$ for the clip- and video-level pretraining, respectively. More hyper-parameter details can be found in Tab.~\ref{tab:hyperparam}.

\subsection{Evaluation Setup.}

We evaluate the representation ability of our OphCLIP using two types of downstream tasks: surgical phase recognition and surgical tool recognition. Additionally, we conduct zero-shot evaluation and linear probing to assess the model's multi-modal alignment and visual representation capabilities. Tables \ref{procedure_abbr}-\ref{caDIS_c_instrument_label} list the specific label names we used for the downstream validation datasets. The labels for the OphNet~\cite{hu2024ophnet} dataset can be found in the online table: \url{https://docs.google.com/spreadsheets/d/1p5lURkth587-lxYwd6eOSmSxPpvIqvyuOKW-4B49PT0/edit?gid=0#gid=0}

\noindent \textbf{Surgical Phase Recognition.} This task evaluates the model's understanding of surgical scenes by classifying video frames into predefined surgical phases. It requires the model to identify instruments, anatomical structures, and their interactions by extracting meaningful visual patterns. To focus on multi-modal representation learning, we exclude temporal modeling and analyze frame-level understanding instead.

\noindent \textbf{Surgical Tool Recognition.} This task tests the model's ability to detect and classify surgical instruments within video frames. By analyzing visual features like shape, texture, and contextual cues, the model demonstrates object-level understanding without reliance on broader workflow context. We assess its robustness in identifying tools despite variations in orientation, scale, or occlusion, emphasizing the quality of learned visual representations.

\begin{table}[ht]
\centering
\resizebox{\columnwidth}{!}{%
\begin{tabular}{l|l}
\toprule[1.1pt]
\textbf{Instrument Label} & \textbf{Textual Prompt} \\ \midrule
Capsulorhexis Forceps & This video shows capsulorhexis forceps. \\
Capsulorhexis Cystotome & This video shows capsulorhexis cystotome. \\
Katena Forceps & This video shows katena forceps. \\
Irrigation-Aspiration & This video shows irrigation aspiration. \\
Slit Knife & This video shows slit knife. \\
Phacoemulsification Tip & This video shows phacoemulsification tip. \\
Spatula & This video shows spatula. \\
Gauge & This video shows gauge. \\
Lens Injector & This video shows lens injector. \\
Incision Knife & This video shows incision knife. \\
\bottomrule
\end{tabular}}
\caption{Textual prompts for each instrument label in the Cataract-1K~\cite{ghamsarian2024cataract} dataset.}
\label{tab:tool_prompt}
\vspace{-0.5cm}
\end{table}

\noindent \textbf{Zero-shot Evaluation.} To perform frame-wise classification tasks for surgical phase and tool recognition, we construct textual prompts tailored to the class labels. For phase recognition, we address their high-level definitions by breaking them down into essential components such as phase, instrument, medication, and goal. These are referred to as keyword-only prompts as shown in Tab.~\ref{tab:phase_prompt}. Additionally, we leverage Large Language Models (LLMs) to generate caption-only prompts, which are detailed descriptive sentences that incorporate relevant surgical instruments, anatomical structures, and events for each phase. These prompts help align the textual domain of pretraining with the downstream task corpus. For tool recognition, we create human-like descriptive sentences to minimize the textual domain gap, ensuring better alignment between pretraining and downstream corpus, as shown in Tab.~\ref{tab:tool_prompt}. This approach facilitates robust zero-shot performance by bridging differences in textual contexts. 

\noindent \textbf{Linear-Probing Evaluation.} For linear-probing, we freeze the visual encoder and train a linear classifier on the extracted features. No image augmentations are applied during training. The linear classifier is implemented as a linear Support Vector Machine (SVM) with a ``linear'' kernel. We fit the model on the training and validation sets, then evaluate its performance on a separate test set. For few-shot linear-probing, we use a $k$-percentage shot approach, tailored for surgical video data. Specifically, we sample 10\% of the videos from the training set, ensuring no data leakage while maintaining a balanced number of samples across classes. This setup allows for a fair evaluation of the model's generalization with limited supervision.

\noindent \textbf{More Ablation Experiments}
Tab.~\ref{tab:ablation2} presents additional results of ablation experiments on the Cataract-101~\cite{schoeffmann2018cataract} and CatRelDet~\citep{hamsarianTPSS20} datasets.

\begin{table}[t!]
\centering
\resizebox{\columnwidth}{!}{%
\begin{tabular}{@{}l|ccc|c@{}}
\toprule
\textbf{Instrument Label} & \textbf{Precision} & \textbf{Recall} & \textbf{F1} & \multicolumn{1}{c}{\textbf{Support}} \\ \midrule
Capsulorhexis Forceps & 6.1 & 100.0 & 11.5 & 100 \\
Capsulorhexis Cystotome & 4.8 & 100.0 & 9.1 & 85 \\
Katena Forceps & 1.6 & 100.0 & 3.1 & 28 \\
Irrigation-Aspiration & 25.4 & 100.0 & 40.5 & 451 \\
Slit Knife & 1.6 & 100.0 & 3.1 & 28 \\
Phacoemulsification Tip & 30.7 & 100.0 & 46.9 & 545 \\
Spatula & 40.3 & 100.0 & 57.4 & 716 \\
Gauge & 24.0 & 100.0 & 38.7 & 426 \\
Lens Injector & 3.7 & 100.0 & 7.2 & 66 \\
Incision Knife & 1.2 & 100.0 & 2.4 & 22 \\ \midrule
Macro Avg. & 13.9 & 100.0 & 22.0 & 2475 \\ \bottomrule
\end{tabular}%
}
\caption{Detailed instrument recognition performance of CLIP~\citep{radford2021learning}, SLIP~\citep{mu2022slip}, and LaCLIP\citep{fan2024improving} on Cat-21 dataset in terms of each class label.}
\label{tab:recall}
\end{table}

\section{Limitation}
\noindent \textbf{Data Bias.} The OphVL dataset is sourced from YouTube, showcasing diverse styles, resolutions, and screen elements. This diversity enhances the evaluation of a model’s generalization ability but may also impact its effectiveness and performance. Some videos in the dataset contain subtitles, watermarks, or additional video windows. Furthermore, regional variability introduces discrepancies in surgical descriptions, such as differences in surgical standards, nomenclature, and definitions influenced by cultural or demographic factors. These characteristics in OphVL reflect the complexity of real-world surgical environments, where ophthalmic microscopes may inherently display various windows or parameters during recording. While these factors pose challenges, they also present opportunities to develop models that are better equipped to handle such diversity.

\noindent \textbf{Downstream Task Limitation.} The zero-shot downstream evaluation datasets for OphCLIP are sourced from publicly available datasets, leveraging their high-quality characteristics and ensuring fair comparisons. However, due to the limited diversity of these datasets—most of which primarily focus on phase recognition and instrument classification in ophthalmology—it is challenging to validate the model on a broader range of vision-language understanding tasks, such as lesion identification or anomaly detection. While the Cataract-1K dataset includes annotations for two types of anomalies, lens rotation and pupil reaction, it does not provide frame-level annotations for these cases.

\begin{table}[t!]
\centering
\scalebox{0.76}{
\begin{tabular}{c|ccc|cc}
\toprule[1.1pt]
{Model}  & {OphVL} & {KB} & {Prompt} & Cataract-101~\cite{schoeffmann2018cataract} & CatRelDet~\citep{hamsarianTPSS20} \\
\hline
CLIP~\citep{radford2021learning} & $\times$ & $\times$ & Caption & 10.0 / 3.3 & 15.3 / 11.9 \\
CLIP~\citep{radford2021learning} & $\checkmark$ & $\times$ & Caption & 36.2 / 25.5 & 26.7 / 23.7 \\
OphCLIP & $\checkmark$ & $\times$ & Caption & 37.1 / 31.9 & 33.6 / 35.4 \\
OphCLIP & $\checkmark$ & $\times$ & Mix  & 31.9 / 28.4 & \textbf{34.5} / \textbf{36.1} \\
OphCLIP & $\checkmark$ & $\checkmark$ & Caption & \textbf{41.1} / \textbf{34.7} & 33.6 / 35.3 \\
OphCLIP & $\checkmark$ & $\checkmark$ & Mix & 39.3 / 33.7 & 32.6 / 34.2 \\
\bottomrule[1.1pt]
\end{tabular}
}
\caption{Ablation study of OphCLIP with various components: OphVL (use of the OphVL pretraining dataset), KB (knowledge base with silent videos), and Prompt (descriptive phase prompts: Caption vs. Mix, which includes additional keywords in the captions). We report Accuracy / F1-score in this table. }
\label{tab:ablation2}
\vspace{-0.5cm}
\end{table}

\begin{table*}[ht]
\centering
\resizebox{\textwidth}{!}{%
\begin{tabular}{@{}l|l|l@{}}
\toprule
\textbf{Phase Label} & \textbf{Caption Only Prompt} & \textbf{Keyword Only Prompt} \\ \midrule
Incision & \begin{tabular}[c]{@{}l@{}}A diamond or steel keratome blade is used to create a small, \\ self-sealing incision in the cornea, providing access to the \\ anterior chamber of the eye. This incision allows the introduction \\ of surgical instruments while maintaining intraocular pressure.\end{tabular} & \begin{tabular}[c]{@{}l@{}}Phase: Initial access; \\ Instrument: Diamond or steel blade; \\ Medication: None; \\ Goal: Create an entry point into the anterior chamber.\end{tabular} \\  \midrule
Viscoelastic & \begin{tabular}[c]{@{}l@{}}A viscoelastic agent, such as sodium hyaluronate, \\ is injected into the anterior chamber using a cannula. \\ This agent maintains space, protects the corneal endothelium, \\ and stabilizes the anterior chamber during the surgery.\end{tabular} & \begin{tabular}[c]{@{}l@{}}Phase: Chamber stabilization; \\ Instrument: Syringe or cannula; \\ Medication: Ophthalmic Viscoelastic Device (OVD); \\ Goal: Maintain anterior chamber depth and protect corneal endothelium.\end{tabular} \\  \midrule
Capsulorhexis & \begin{tabular}[c]{@{}l@{}}Using capsulorhexis forceps or a cystotome, \\ the surgeon creates a circular tear in the anterior lens capsule. \\ This opening allows access to the underlying cataractous lens, \\ preparing it for removal.\end{tabular} & \begin{tabular}[c]{@{}l@{}}Phase: Capsule opening; \\ Instrument: Forceps or cystotome; \\ Medication: None (Viscoelastic used for support); \\ Goal: Create a circular opening in the anterior lens capsule.\end{tabular} \\  \midrule
Hydrodissection & \begin{tabular}[c]{@{}l@{}}Balanced salt solution (BSS) is injected with a cannula between \\ the lens capsule and the lens cortex, separating the cataract from \\ the capsule. This ensures that the lens material can be removed more \\ easily during phacoemulsification.\end{tabular} & \begin{tabular}[c]{@{}l@{}}Phase: Lens loosening; \\ Instrument: Cannula; \\ Medication: Balanced Salt Solution (BSS); \\ Goal: Separate the lens cortex from the capsule for easy extraction.\end{tabular} \\  \midrule
Phacoemulsification & \begin{tabular}[c]{@{}l@{}}A phacoemulsification handpiece with an ultrasonic probe is \\ inserted into the eye to emulsify the cataract into tiny fragments. \\ These fragments are simultaneously aspirated, removing the \\ clouded lens while protecting surrounding structures.\end{tabular} & \begin{tabular}[c]{@{}l@{}}Phase: Lens removal; Instrument: \\ Phacoemulsification handpiece; \\ Medication: Balanced Salt Solution (BSS) for cooling and irrigation; \\ Goal: Break up and emulsify the cataract for extraction.\end{tabular} \\ \midrule
Irrigation/Aspiration & \begin{tabular}[c]{@{}l@{}}A dual-function irrigation and aspiration (I/A) handpiece is used \\ to remove any remaining lens material and fluid from the \\ capsular bag and anterior chamber. The procedure ensures the \\ capsular bag is clear for lens implantation.\end{tabular} & \begin{tabular}[c]{@{}l@{}}Phase: Lens material removal; \\ Instrument: Irrigation/Aspiration handpiece; \\ Medication: Balanced Salt Solution (BSS); \\ Goal: Remove remaining lens fragments from the capsular bag.\end{tabular} \\ \midrule
Capsule Pulishing & \begin{tabular}[c]{@{}l@{}}A polishing tip or I/A tool is used to gently remove residual \\ epithelial cells from the inner surface of the posterior capsule, \\ minimizing the risk of posterior capsule opacification\\ (secondary cataract formation).\end{tabular} & \begin{tabular}[c]{@{}l@{}}Phase: Capsule cleaning; \\ Instrument: Polishing tip or Irrigation/Aspiration tool; \\ Medication: None; \\ Goal: Remove residual lens epithelial cells to reduce posterior capsule opacification.\end{tabular} \\ \midrule
Lens Implantation & \begin{tabular}[c]{@{}l@{}}An intraocular lens (IOL) is loaded into an injector and inserted \\ through the corneal incision. It is placed within the capsular bag \\ to replace the natural lens and restore the patient’s vision.\end{tabular} & \begin{tabular}[c]{@{}l@{}}Phase: Lens insertion; \\ Instrument: Intraocular lens (IOL) injector; \\ Medication: None; \\ Goal: Insert the artificial intraocular lens into the capsular bag.\end{tabular} \\ \midrule
Lens positioning & \begin{tabular}[c]{@{}l@{}}Using fine-tipped instruments, the surgeon carefully adjusts the \\ position of the IOL within the capsular bag to ensure proper \\ centration and stability, optimizing visual outcomes.\end{tabular} & \begin{tabular}[c]{@{}l@{}}Phase: Lens alignment; \\ Instrument: Manipulating hook or forceps; \\ Medication: None; \\ Goal: Ensure the intraocular lens is correctly positioned and centered.\end{tabular} \\ \midrule
Viscoelastic\_Suction & \begin{tabular}[c]{@{}l@{}}The viscoelastic agent is aspirated from the anterior chamber \\ using the I/A handpiece to prevent postoperative pressure spikes \\ and ensure a clear visual axis.\end{tabular} & \begin{tabular}[c]{@{}l@{}}Phase: Viscoelastic removal; \\ Instrument: Irrigation/Aspiration handpiece; \\ Medication: None; \\ Goal: Remove any remaining viscoelastic agents from the anterior chamber.\end{tabular} \\ \midrule
Anterior\_Chamber Flushing & \begin{tabular}[c]{@{}l@{}}The anterior chamber is flushed with balanced salt solution (BSS) \\ to remove any remaining debris or blood. This final rinse ensures \\ that the chamber is clear and that the incision site is clean.\end{tabular} & \begin{tabular}[c]{@{}l@{}}Phase: Final chamber cleaning; \\ Instrument: Irrigation/Aspiration handpiece; \\ Medication: Balanced Salt Solution (BSS); \\ Goal: Ensure the anterior chamber is clear of any debris or substances.\end{tabular} \\ \midrule
Tonifying/Antibiotics & \begin{tabular}[c]{@{}l@{}}A pupil-constricting agent, such as acetylcholine, may be injected \\ to stabilize intraocular pressure. Following this, \\ an antibiotic such as moxifloxacin is administered to prevent infection, \\ and sometimes corticosteroids are used to reduce inflammation.\end{tabular} & \begin{tabular}[c]{@{}l@{}}Phase: Final stabilization and protection; \\ Instrument: Syringe or cannula; \\ Medication: Acetylcholine (for pupil constriction) and moxifloxacin (antibiotic); \\ Goal: Stabilize intraocular pressure and prevent infection.\end{tabular} \\ \bottomrule
\end{tabular}%
}
\caption{\textbf{Prompt example.} Caption-only and keyword-only prompts for each phase label in the Cataract-1K~\cite{ghamsarian2024cataract} dataset, respectively.}
\label{tab:phase_prompt}
\end{table*}
\clearpage

\begin{figure*}[t!]
\centering
\begin{subfigure}[t]{\linewidth}
    \centering
    \includegraphics[width=0.95\linewidth]{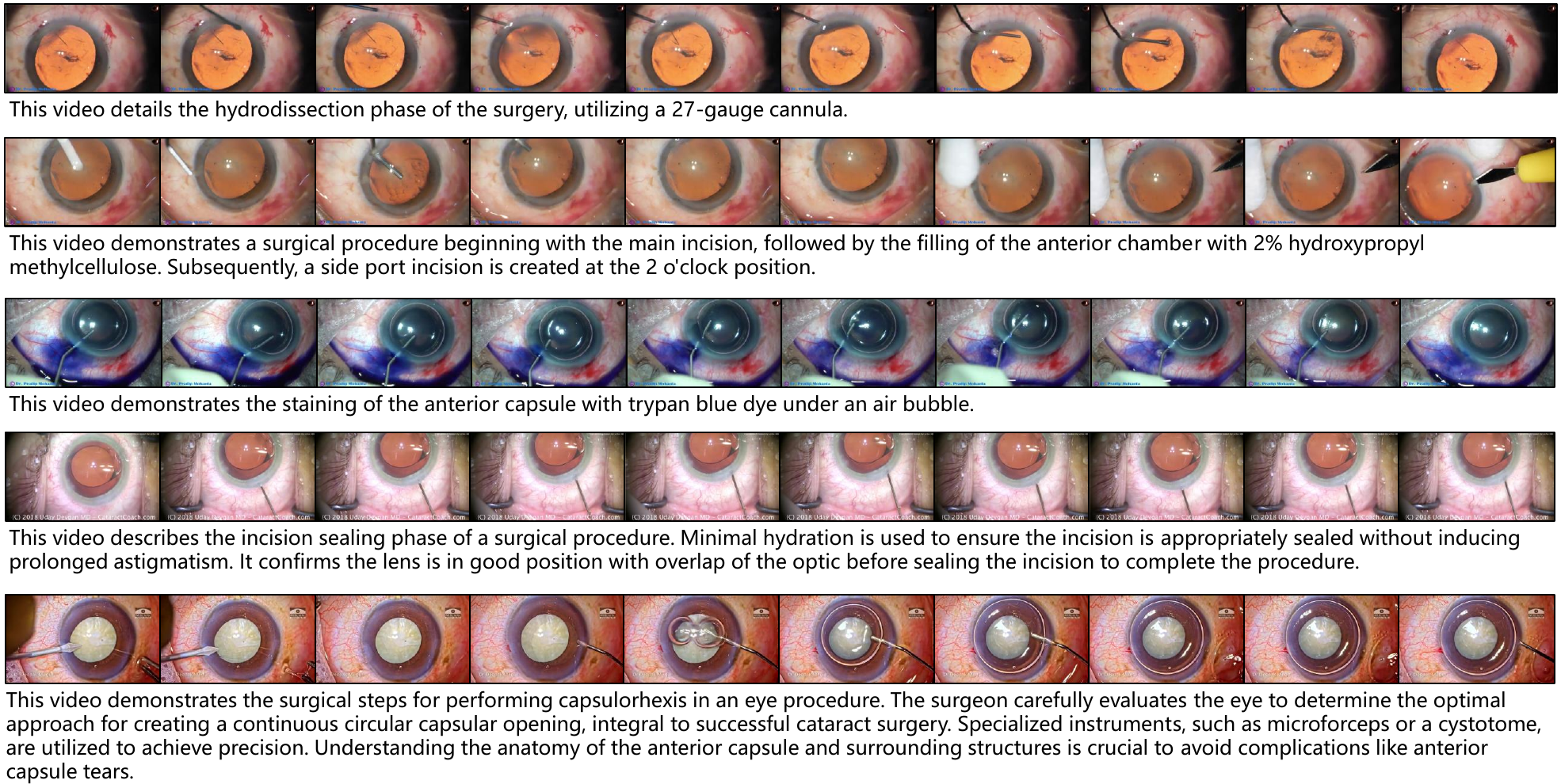}
\end{subfigure}

\begin{subfigure}[t]{\linewidth}
    \centering
    \includegraphics[width=0.95\linewidth]{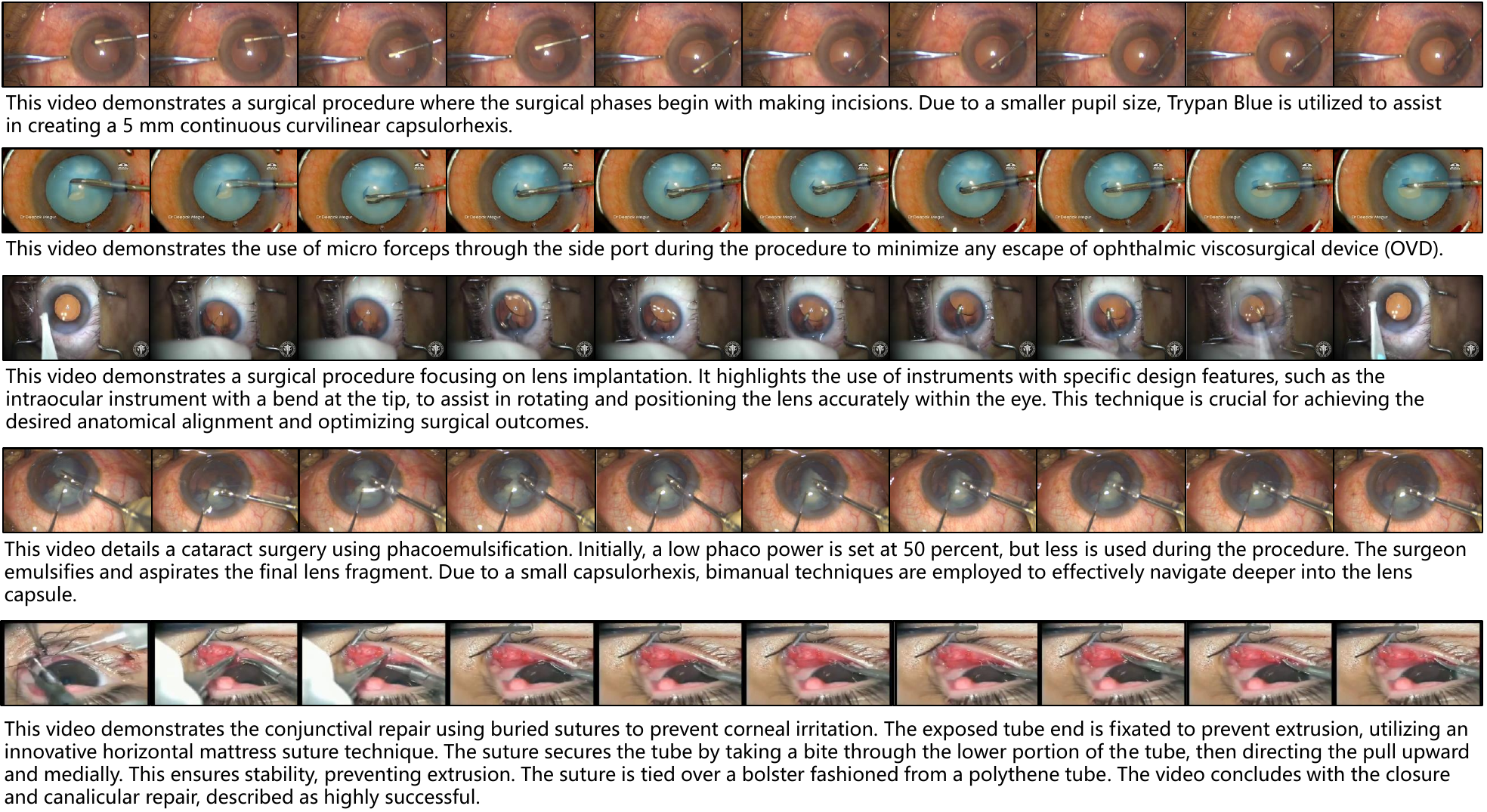} 
\end{subfigure}

\label{pair_sample}
\end{figure*}

\begin{figure*}[t!]
\centering
\begin{subfigure}[t]{\linewidth}
    \centering
    \includegraphics[width=0.95\linewidth]{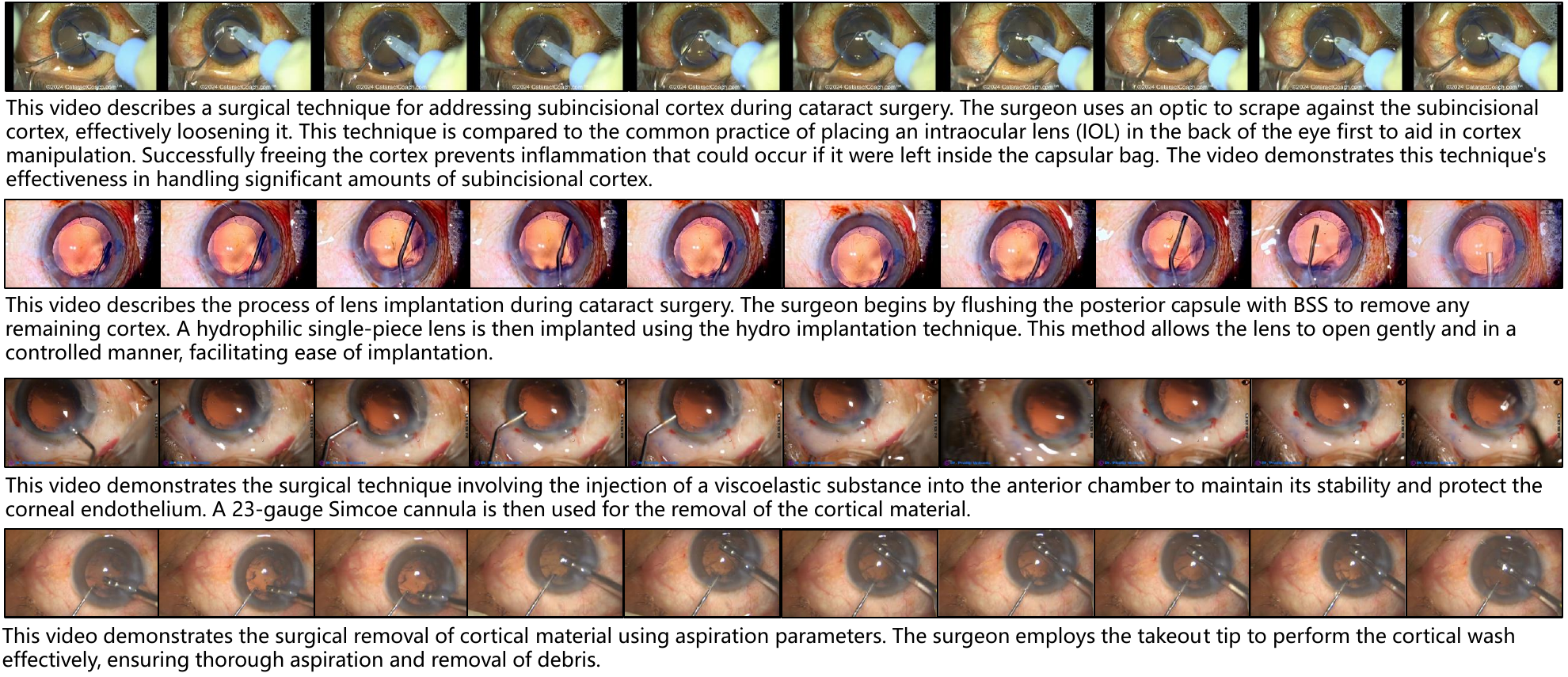}
\end{subfigure}

\begin{subfigure}[t]{\linewidth}
    \centering
    \includegraphics[width=0.95\linewidth]{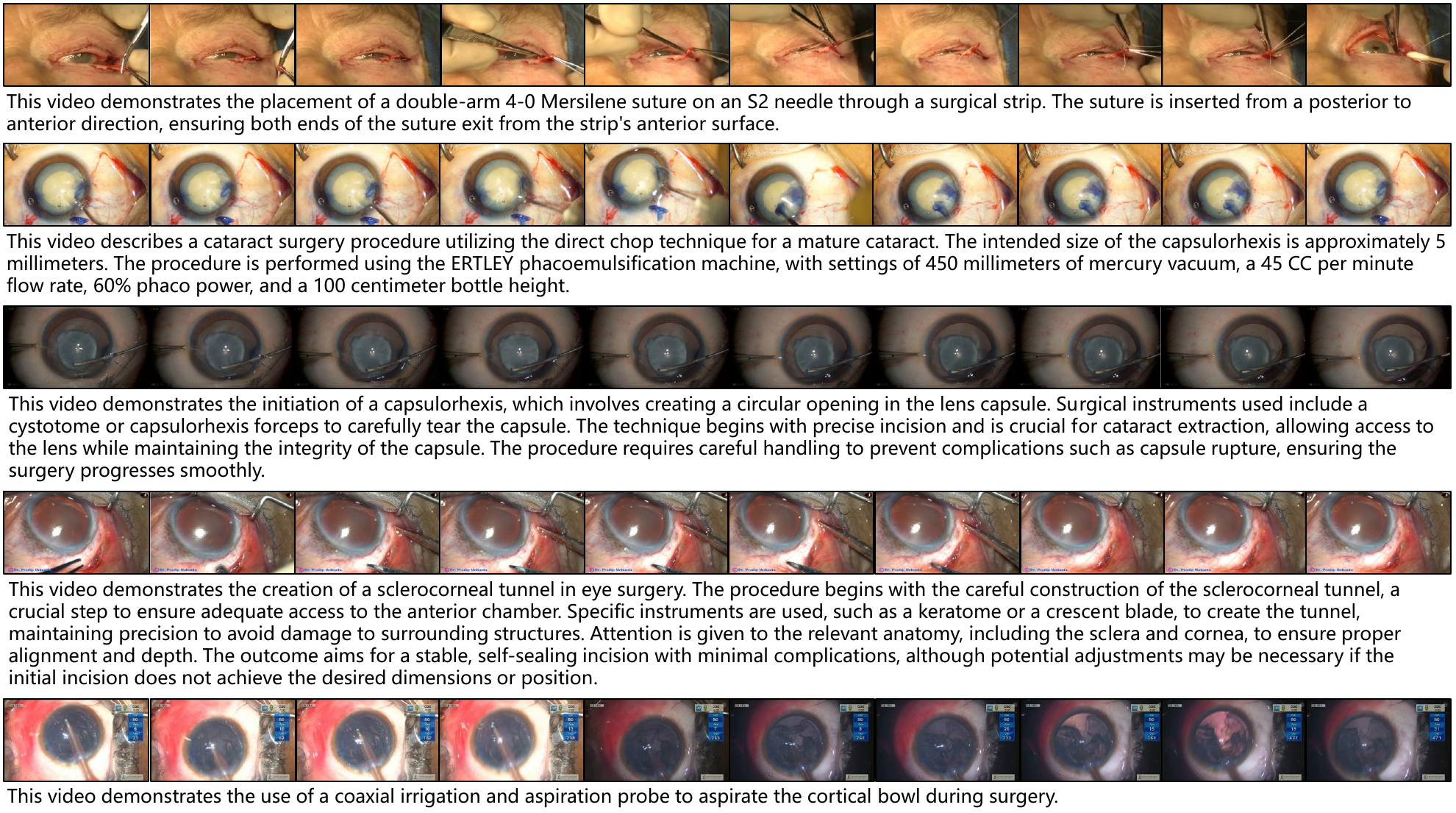}
\end{subfigure}

\caption{\textbf{Some examples of clip-text pairs from OphVL.}}
\label{pair_sample}
\end{figure*}

\clearpage

\begin{table}[t!]
\centering
\begin{tabularx}{\columnwidth}{ll}
\toprule
\textbf{ID}  & \textbf{Phase}\\
\midrule
0 & Antibiotikum \\
1 & Hydrodissektion \\
2 & Incision \\
3 & Irrigation-Aspiration \\
4 & Kapselpolishing \\
5 & Linsenimplantation \\
6 & Phako \\
7 & Rhexis \\
8 & Tonisieren \\
9 & Visco-Absaugung \\
10 & Viscoelasticum \\
11 & not\_initialized \\
\bottomrule
\end{tabularx} 
\caption{Phase labels of the Cat-21~\citep{PrimusPTMEBS18} dataset.}
\label{procedure_abbr}
\end{table}

\begin{table}[t!]
\centering
\begin{tabularx}{\columnwidth}{ll}
\toprule
\textbf{ID}  & \textbf{Phase}\\
\midrule
0 & Incision \\
1 & Viscoelastic \\
2 & Capsulorhexis \\
3 & Hydrodissection \\
4 & Phacoemulsification \\
5 & Irrigation/Aspiration \\
6 & Capsule Pulishing \\
7 & Lens Implantation \\
8 & Lens positioning \\
9 & Viscoelastic\_Suction \\
10 & Anterior\_Chamber Flushing \\
11 & Tonifying/Antibiotics \\
\bottomrule
\end{tabularx} 
\caption{Phase labels of the Cataract-1K~\citep{ghamsarian2024cataract} dataset.}
\label{catarack-1k_phase_label}
\end{table}

\begin{table}[t!]
\centering
\begin{tabularx}{\columnwidth}{ll}
\toprule
\textbf{ID}  & \textbf{Phase} \\
\midrule
0 & Incision \\
1 & Viscous agent injection \\
2 & Rhexis \\
3 & Hydrodissection \\
4 & Phacoemulsificiation \\
5 & Irrigation and aspiration \\
6 & Capsule polishing \\
7 & Lens implant setting-up \\
8 & Viscous agent removal \\
9 & Tonifying and antibiotics \\
\bottomrule
\end{tabularx} 
\caption{Phase labels of the Cataract-101~\citep{schoeffmann2018cataract} dataset.}
\label{catarack-101_phase_label}
\end{table}

\begin{table}[t!]
\centering
\begin{tabularx}{\columnwidth}{ll}
\toprule
\textbf{ID} & \textbf{Phase} \\
\midrule
0 & Implantation \\
1 & Irrigation\_Aspiration and Visc\_Suction \\
2 & Phacoemulsification \\
3 & Rhexis \\
4 & Rest \\
\bottomrule
\end{tabularx} 
\caption{Phase labels of the CatRelDet~\citep{hamsarianTPSS20} dataset.}
\label{catreldet_phase_label}
\end{table}

\begin{table}[t!]
\centering
\begin{tabularx}{\columnwidth}{ll}
\toprule
\textbf{ID}  & \textbf{Phase} \\
\midrule
0 & Linsenimplantation \\
1 & Linsenimplantation\_before \\
2 & Linsenimplantation\_after \\
\bottomrule
\end{tabularx} 
\caption{Phase labels of the LensID~\citep{GhamsarianTPSES21} dataset.}
\label{lensID_label}
\end{table}

\begin{table}[t!]
\centering
\begin{tabularx}{\columnwidth}{ll}
\toprule
\textbf{ID}  & \textbf{Instrument} \\
\midrule
0 & spatula \\
1 & 27 gauge cannula \\
2 & slit knife \\
3 & phaco tip \\
4 & capsulorhexis forceps \\
5 & cartridge \\
6 & I/A handpiece \\
7 & cannula \\
8 & katena forceps \\
9 & eye retractors \\
10 & angled incision knife \\
\bottomrule
\end{tabularx} 
\caption{Instrument labels of the CatInstSeg~\citep{FoxTS20} dataset.}
\label{insegcat_instrument_label}
\end{table}

\begin{table}[t!]
\centering
\begin{tabularx}{\columnwidth}{ll}
\toprule
\textbf{ID}  & \textbf{Instrument} \\
\midrule
0 & Capsulorhexis Forceps \\ 
1 & Capsulorhexis Cystotome \\
2 & Katena Forceps \\
3 & Irrigation-Aspiration \\
4 & Slit Knife \\
5 & Phacoemulsification Tip \\
6 & Spatula \\
7 & Gauge \\
8 & Lens Injector \\
9 & Incision Knife \\
\bottomrule
\end{tabularx} 
\caption{Instrument labels of the Cataract-1K~\citep{ghamsarian2024cataract} dataset.}
\label{cataract-1k_instrument_label}
\end{table}

\begin{table}[t!]
\centering
\begin{tabularx}{\columnwidth}{ll}
\toprule
\textbf{ID}  & \textbf{Instrument} \\
\midrule
0 & I/A Handpiece \\
1 & Marker \\
2 & Rycroft Cannula Handle \\
3 & Eye Retractors \\
4 & Cotton \\
5 & Secondary Knife Handle \\
6 & Surgical Tape \\
7 & Troutman Forceps \\
8 & Hydrodissection Cannula Handle \\
9 & Vitrectomy Handpiece \\
10 & Iris Hooks \\
11 & Rycroft Cannula \\
12 & Lens Injector \\
13 & Secondary Knife \\
14 & Mendez Ring \\
15 & Primary Knife \\
16 & Capsulorhexis Cystotome \\
17 & I/A Handpiece Handle \\
18 & Micromanipulator \\
19 & Charleux Cannula \\
20 & Phacoemulsifier Handpiece \\
21 & Viscoelastic Cannula \\
22 & Capsulorhexis Forceps \\
23 & Phacoemulsifier Handpiece Handle \\
24 & Lens Injector Handle \\
25 & background \\
26 & Hydrodissection Cannula \\
27 & Capsulorhexis Cystotome Handle \\
28 & Needle Holder \\
29 & Suture Needle \\
30 & Bonn Forceps \\
31 & Primary Knife Handle \\
\bottomrule
\end{tabularx} 
\caption{Fine-grained instrument labels of the CaDIS~\cite{grammatikopoulou2021cadis} dataset (CaDIS-F).}
\label{caDIS_instrument_label}
\end{table}

\begin{table}[t!]
\centering
\begin{tabularx}{\columnwidth}{ll}
\toprule
\textbf{ID}  & \textbf{Instrument} \\
\midrule
0 & I/A Handpiece \\
1 & Cap. Forceps \\
2& Eye Retractors \\
3 & Lens Injector \\
4 & Tissue Forceps \\
5 & Surgical Tape \\
6 & Ph. Handpiece \\
7 & Cannula \\
8 & Secondary Knife \\
9 & Cap. Cystotome \\
10 & Primary Knife \\
11 & Micromanipulator \\
\bottomrule
\end{tabularx} 
\caption{Coarse-grained instrument labels of the CaDIS~\cite{grammatikopoulou2021cadis} dataset (CaDIS-C).}
\label{caDIS_c_instrument_label}
\end{table}

\clearpage

{
    \small
    \bibliographystyle{ieeenat_fullname}
    \bibliography{main}
}


\end{document}